\pdfoutput=1

\documentclass[11pt]{article}

\usepackage[final]{acl}

\usepackage{times}
\usepackage{latexsym}

\usepackage[T1]{fontenc}

\usepackage[utf8]{inputenc}

\usepackage{microtype}

\usepackage{inconsolata}

\usepackage{graphicx}

\usepackage{amsmath}
\usepackage{amssymb}
\usepackage{caption}
\usepackage{subcaption}
\usepackage{algorithm}
\usepackage{algpseudocode}

\DeclareMathOperator{\pa}{\operatorname{pa}}
\DeclareMathOperator{\argmin}{\operatorname{argmin}}

\title{Generating Synthetic Relational Tabular Data via
Structural Causal Models}

 \author{Frederik Hoppe \and Astrid Franz \and Lars Kleinemeier \and Udo G\"obel \\
         CONTACT Software GmbH, Wiener Str.~1-3, 28359 Bremen, Germany \\ \texttt{frederik.hoppe@contact-software.com} }

\begin{document}
\maketitle
\begin{abstract}
Synthetic tabular data generation has received increasing attention in recent years, particularly with the emergence of foundation models for tabular data. The breakthrough success of TabPFN \cite{hollmann2025tabpfn}, which leverages vast quantities of synthetic tabular datasets derived from structural causal models (SCMs), demonstrates the critical role synthetic data plays in developing powerful tabular foundation models. However, most real-world tabular data exists in relational formats spanning multiple interconnected tables — a structure not adequately addressed by current generation methods. In this work, we extend the SCM-based approach by developing a novel framework that generates realistic synthetic relational tabular data including causal relationships across tables. Our experiments confirm that this framework is able to construct relational datasets with complex inter-table dependencies mimicking real-world scenarios.
\end{abstract}

\section{Introduction}
The development of synthetic data generation techniques has seen remarkable progress with the advent of foundation models, particularly in domains such as images and text. However, generating realistic tabular data - especially relational tabular data with properly linked entries - remains an underexplored challenge in machine learning research. While large language models and diffusion models have revolutionized synthetic data generation across various domains, structured tabular data has received comparatively less attention despite its prevalence in real-world applications.

In this paper, we develop a novel synthetic relational data generation framework for creating arbitrarily large amounts of relational datasets with complex, realistic dependencies, suitable, e.g., for foundation model training and benchmark creation. In order to systematically model both intra-table correlations and inter-table relationships, our method constructs data independently from real-world datasets, overcoming accessibility limitations. Our method is inspired by the SCM-based approach for single tables of TabPFN \cite{hollmann2025tabpfn}. However, we introduce critical changes to the original SCM framework, and extend it to generate multiple tables connected through shared key columns. Based on these extensions, we provide an automated framework for creating synthetic relational datasets that comprise both statistical properties within individual tables and structural relationships between them. This contribution enables the creation of realistic relational tabular data that can be used for developing models capturing inter-table relationships.

\section{Related Work}
Synthetic tabular data generation has evolved significantly to address challenges like data scarcity and privacy concerns. 
Earlier work \cite{patkiSyntheticDataVault2016} presented the Synthetic Data Vault, which builds generative models of relational databases through recursive conditional parameter aggregation. It is the first learning-based approach for generating relational data.
Recently, \cite{hudovernikRelationalDataGeneration} proposed an approach that combines graph neural network embeddings with diffusion models, exploiting a graph representation of relational data induced by foreign key constraints. The method captures topological structure and statistical properties across multiple linked tables. 
These approaches require a (real-world) dataset as a basis to extract statistical and relational patterns, which are then used to generate new data with the same statistical properties. However, due to the lack of accessible real-world datasets, these methods seem to be unsuited for producing huge amounts of relational datasets with manifold intra- and inter-table relationships.
A generation method independent of real-world data was proposed in \cite{hollmann2025tabpfn}, generating synthetic datasets through SCMs \cite{Pearl2010}, which naturally allow for simulating wide-ranging causal dependencies. However, the method is restricted to single, unrelated tables.

\section{Data Generation Method\label{sec:method}}
Our structured data generation approach is based on an SCM, represented by a directed acyclic graph (DAG) $\mathcal{G}$ with directed edges, connecting parent nodes (causes) to child nodes (effects). For every node $i=1,\hdots, N$, a structural assignment
\begin{equation}\label{eq:scm_equation}
	x_i = f_i(x_{\pa(i)}, \varepsilon_i)\in\mathbb{R}^n
\end{equation}
is used to propagate the data in $\mathcal{G}$, where $n$ denotes the hidden dimension of the data at each node, $f_i$ a deterministic mapping, $x_{\pa(i)}$ the realization of the parent data of node $i$, and $\varepsilon_i$ an independent $n$-dimensional noise vector.
First, we sample the structure of the model, i.e.\ the nodes and the directed edges. Second, for every independent sample, i.e.\ table row, we initialize the data as multi-dimensional vectors at the root nodes and propagate it, including random noise, through the graph. In the final step, we readout the data by projecting the $n$-dimensional vectors to scalars. Thus, we obtain a two-dimensional data scheme where the number of rows corresponds to the number of samples and the number of columns to the number of nodes. Subsections~\ref{subsec:structure_sampling}, \ref{subsec:data_sampling} and \ref{subsec:data_readout} describe these construction steps in more detail. Algorithm~\ref{alg:dataset_generation} provides a high-level overview.

After these steps, the data of the final tabular format could be additionally processed via bias induction, disturbance by additional noise, wrong data incorporation or the methods mentioned in \cite{hollmann2025tabpfn} in order to mimic real-world data challenges.

\begin{algorithm}
\caption{Generating Synthetic Datasets}\label{alg:dataset_generation}
\begin{algorithmic}
\State \hspace*{-0.4cm}\textbf{Structure Sampling:} \Comment{cf. Subsec.~\ref{subsec:structure_sampling}}
\State Sample DAG $\mathcal{G}$
\State Initialize root node distributions
\State \hspace*{0.4cm}Define propagation fct $g_i$ (e.g. neural net)
\State \hspace*{0.4cm}Define pooling fct $p_i$ (e.g. norm, categorical)
\State \hspace*{-0.4cm}\textbf{Pre-Sampling:} \Comment{cf. Subsec.~\ref{subsec:data_sampling}}
\State Sample root data and propagate it by $g_i$
\State {\bfseries for} every node $i$ {\bfseries do}
\State \hspace*{0.4cm}Compute component-wise 10\%- and 90\%-
\State \hspace*{0.4cm}quantiles $q_{0.1}(i),q_{0.9}(i)$
\State \hspace*{0.4cm}{\bfseries if} $i$ categorical node
\State \hspace*{0.8cm}Choose number of categories $K$
\State \hspace*{0.8cm}Cluster data into $K$ categories
\State \hspace*{0.8cm}Refine $p_i$ as in \eqref{eq:pooling_cat}
\State \hspace*{-0.4cm}\textbf{Main Data Sampling:} \Comment{cf. Subsec.~\ref{subsec:data_readout}}
\State Sample root data and propagate it by $f_i$ (cf.~\eqref{eq:noise_propagation})
\State Read out data at every node by $p_i$
\end{algorithmic}
\end{algorithm}

\subsection{Structure Sampling}\label{subsec:structure_sampling}
To sample a directed graph, we utilize the Barabási Albert model \cite{BarabasiAlbertGraph}. After removing isolated nodes and edges $(i,j)$ with $j>i$, we obtain a DAG. The sinks of the resulting graph represent the targets, and the remaining nodes are considered as features for the future dataset. 

The data associated with the root nodes are initialized as $n$-dimensional vectors, drawn from a range of distributions, including normal distributions with random means and standard deviations as well as gamma distributions with random shapes and scales.
This initialization data will be propagated through the graph.
For every node $i=1,\hdots, N$, we define a function 
\begin{equation}\label{eq:scm_propagate}
    g_i:\mathbb{R}^{\vert \pa(i)\vert\cdot n}\to\mathbb{R}^n 
\end{equation} 
that propagates the concatenated parent data. In contrast to \cite{hollmann2025tabpfn}, we do not incorporate categorical feature generation into the set of propagation functions (\ref{eq:scm_propagate}), since this restricts the variety of states at successive nodes. In our approach, the data is propagated through the graph as multi-dimensional (continuous) vectors. Only when we observe the data, it may become categorical, i.e.\ discretized. Thus, we construct categorical data without restricting the number of different data vectors at the subsequent nodes to the number of categories. In principle, propagation functions (\ref{eq:scm_propagate}) could be arbitrarily defined. In this work, propagation functions (\ref{eq:scm_propagate}) are considered to be one-layer fully-connected neural networks followed by randomly chosen (non-)linear activation functions, e.g., ReLU, logabs.

Once each node has processed the data, the information from the resulting vectors $x_1,\hdots, x_N$ is stored in a tabular format. To this end, we define a set of pooling functions $p_i:\mathbb{R}^n \to \mathbb{R}$, to reduce dimensionality, i.e.,
for every node $i=1,\hdots, N$, we independently select a continuous pooling function $p_i$ such as norm, mean, median or variance of the vector, or a categorical pooling function, defined in Subsection~\ref{subsec:data_sampling}.

\begin{figure}
	\includegraphics[width=\linewidth]{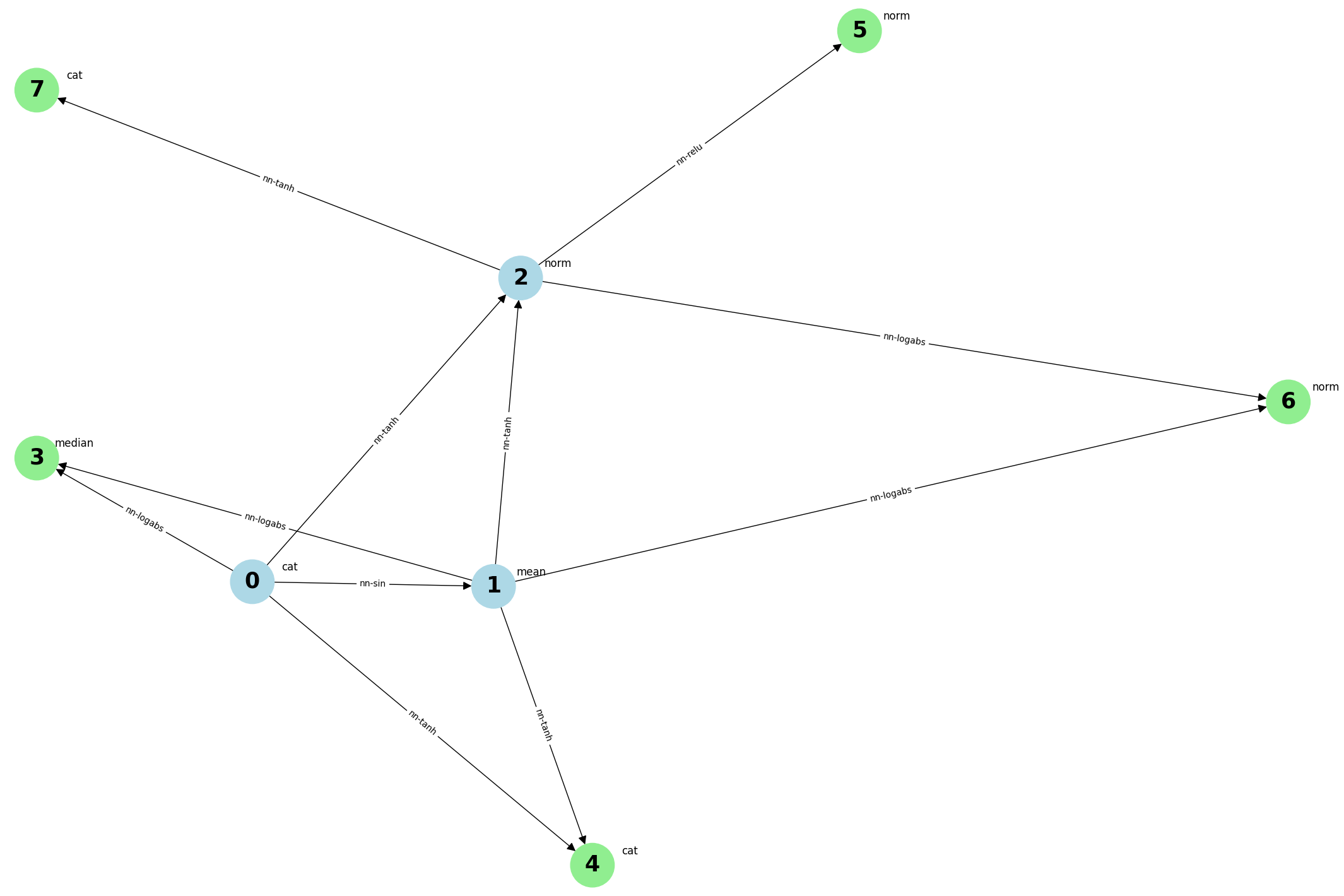}
	\caption{Example of a DAG illustrating the SCM. Nodes represent the structural assignments, see Equation~\eqref{eq:scm_equation}, annotated with the corresponding pooling function (Euclidean {\bfseries norm}, {\bfseries mean}, {\bfseries median} or {\bfseries cat}egorical projection). Edges indicate the flow of the data vectors, with edge labels specifying the applied (non)-linear activation function. The green nodes symbolize the targets, while the blue nodes correspond to the features.
	\label{fig:dag}}
\end{figure}

The selection of initialization and pooling functions is stored in the graph structure to assure reproducability and allow for a detailed analysis of parameter influence. An example of such a DAG is presented in Fig.~\ref{fig:dag}.

\subsection{Pre-Sampling\label{subsec:data_sampling}}
Propagating the data in the DAG $\mathcal{G}$ according to Equation~\eqref{eq:scm_equation} involves noise vectors $\varepsilon_i$. In order to align the magnitude of the noise influence with the data distribution, we conduct the data generation process via a low-sample pre-run without noise. We independently sample the root data according to the pre-defined distribution and propagate the data through the entire graph. Then, we estimate the corresponding data distribution of each node with the sampling data of the pre-run. More concretely, for every node $i$ we compute the $10\%$- and $90\%$-quantiles component-wise, denoted by vectors $q_{0.1}$ and $q_{0.9}$. In the main data generation run, this information enables us to tailor the noise level to the distribution of the corresponding node, i.e., $x_i$ is given by 
\begin{align}
	x_i &= f_i(x_{\pa(i)}, \varepsilon_i) \nonumber\\
	&= g_i(x_{\pa(i)}) + ( q_{0.9}(i) - q_{0.1}(i)) \varepsilon_i.\label{eq:noise_propagation}
\end{align}

By introducing this noise scaling we ensure a balanced noise integration into the data vector, such that $g_i(x_{\pa(i)})$ remains the primary source of information. The degree of perturbation could be adjusted by computing different quantiles. 

Moreover, the estimation of the node distributions allows for a semantically meaningful discretization of the data into categories. For every categorical node $i$, we randomly select the number of categories $K(i)$, and cluster the pre-sampled data into these $K(i)$ categories, for instance by the k-means algorithm. Then, we define a categorical pooling function, assigning the continuous $n$-dimensional data vectors to the categories:
\begin{equation}\label{eq:pooling_cat}
	p_i(x_i)  = \argmin\limits_{l=1,\hdots, K(i)} \Vert x_i - v_l \Vert_2,
\end{equation}
where $v_1,\ldots,v_{K(i)}$ denote the cluster centroids. It is important to note that the categorization of the data occurs only for the readout and does not affect the subsequent propagation through the child nodes. 

With the information collected by the pre-run, we are able to perform the main sampling run.

\subsection{Main Data Sampling\label{subsec:data_readout}}
During the main run, we initialize independently data at the root nodes and propagate it through the entire graph according to Equations~\eqref{eq:scm_equation} and \eqref{eq:noise_propagation}. Utilizing the pooling functions mentioned above, we project the $n$-dimensional vector at each node to a scalar, which yields one data sample. The procedure is repeated for the desired sample size, resulting in a table, where the rows correspond to the samples and the columns to the graph nodes.

We consider the columns, represented by the sinks of the graph (colored green in Fig.~\ref{fig:dag}), as potential targets whereas the remaining columns (indicated blue in Fig.~\ref{fig:dag}) are considered as features. On the one hand, this allows for a complete usage of the dataset, e.g., to train a tabular foundation model. On the other hand, the single targets could be used independently, e.g., for handling end-to-end scenarios.

\section{Extensions to Relational Data\label{sec:relational}}

\begin{algorithm}[b]
\caption{Generating Relational Datasets}\label{alg:rel_dataset_generation}
\begin{algorithmic}
\State \hspace*{-0.4cm}Sample DAGs $\mathcal{G}_{\text{main}}$ and $\mathcal{G}_{\text{add}}$
\State \hspace*{-0.4cm}Connect via coupling node C: $\mathcal{G}_{\text{add}}\rightarrow \text{C}\rightarrow \mathcal{G}_{\text{main}}$
\State \hspace*{-0.4cm}Link feature nodes of $\mathcal{G}_{\text{add}}$ to target nodes of $\mathcal{G}_{\text{main}}$
\State \hspace*{-0.4cm}Sample dataset w.r.t. Algorithm~\ref{alg:dataset_generation} for merged graph
\State \hspace*{-0.4cm}Sample dataset w.r.t. Algorithm~\ref{alg:dataset_generation} for $\mathcal{G}_{\text{add}}$ (incl. C)
\end{algorithmic}
\end{algorithm}

This section extends the previously described methodology to generate relational tables, summarized in Algorithm~\ref{alg:rel_dataset_generation}. The objective is to create two tables with different sample sizes that share a common feature, represented as a coupling node. First, we independently sample two DAGs, denoted by $\mathcal{G}_{\text{main}}$ and $\mathcal{G}_{\text{add}}$, according to the procedure described in Subsection \ref{subsec:structure_sampling}. Then, we introduce a coupling node C that is caused by a sink of $\mathcal{G}_{\text{add}}$ and directs to a feature of $\mathcal{G}_{\text{main}}$. In this way, we establish a relationship between these graphs and assure, that information propagates from $\mathcal{G}_{\text{add}}$ to $\mathcal{G}_{\text{main}}$. 

\begin{figure}[h]
	\includegraphics[width=\linewidth]{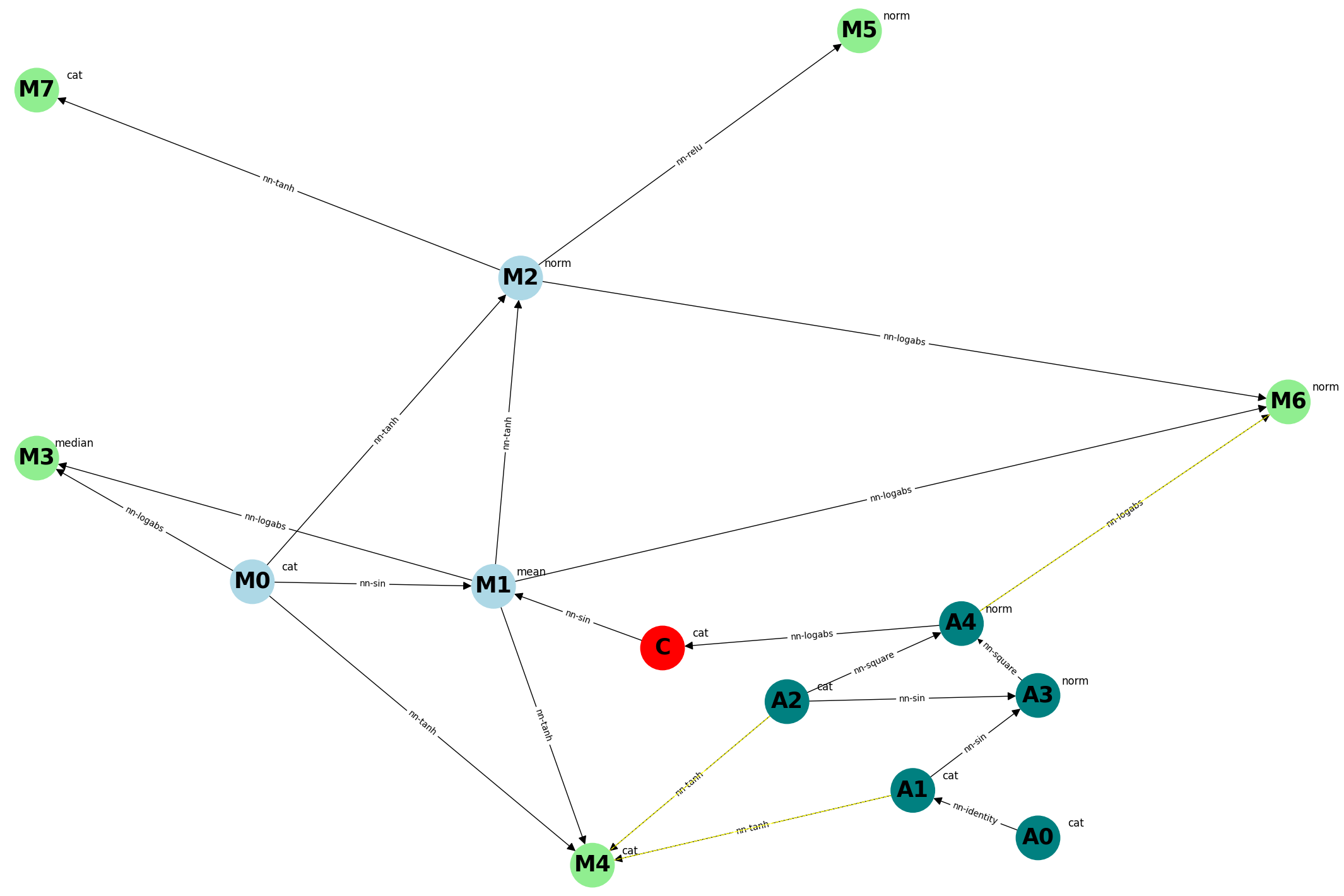}
	\caption{Example graph for generating relational tables. The DAG from Fig.~\ref{fig:dag} with renamed nodes M1, M2, $\ldots$ is extended by an additional DAG with feature nodes A1, A2, $\ldots$. Both graphs are linked via the connection node C and latent relationships indicated by yellow edges.
	\label{fig:combined_graph}}
\end{figure}

To incorporate \emph{latent} causal influence from $\mathcal{G}_{\text{add}}$ to $\mathcal{G}_{\text{main}}$, we connect feature nodes of $\mathcal{G}_{\text{add}}$ to target nodes of $\mathcal{G}_{\text{main}}$. An example of two coupled graphs is illustrated in Fig.~\ref{fig:combined_graph}. Nodes belonging to the main graph $\mathcal{G}_{\text{main}}$ are labeled M1, M2, $\ldots$, forming the same graph as shown in Fig.~\ref{fig:dag}. The nodes in the additional graph $\mathcal{G}_{\text{add}}$ are denoted by A1, A2, $\ldots$. In the same way, more than two relational tables can be generated. The data samples are generated once for the merged graph (including node C) and once for graph $\mathcal{G}_{\text{add}}$ (including node C), both with separate sample sizes. Although we utilize the merged graph to generate the data together with the (latent) causal relationships, the main table contains only the data corresponding to the nodes of $\mathcal{G}_{\text{main}}$ (including node C). The headers of the resulting main and additional tables for the example graphs of Fig.~\ref{fig:combined_graph} are presented in Tables~\ref{tab:main_table} and \ref{tab:add_table}. The edges representing the latent causality effectively link the two tables. Consequently, a comprehensive understanding of the affected targets requires integrating information from both tables, as demonstrated in Section \ref{sec:experiments}. Without the latent causality links, the information propagated through node C would be sufficient to represent the relationships in $\mathcal{G}_{\text{main}}$.

\section{Evaluation\label{sec:experiments}}
We exemplary analyze one relational dataset consisting of two coupled tables constructed as stated in Section~\ref{sec:relational}. This example dataset is based on the DAG shown in Fig.~\ref{fig:combined_graph}. We sample the main dataset with 100,000 rows, described and illustrated in Appendix~\ref{sec:appendix} and Table~\ref{tab:main_table}. The first 90,000 rows serve as training set, while the remaining 10,000 rows are excluded from embedding training and serve as a test set. The additional dataset, including the C-column, is sampled with 500 rows, see Table~\ref{tab:add_table}. In order to measure how the data in the main table is influenced by the data in the additional table, we perform the classical ML tasks classification and regression, first using the main table only and second using the data of the additional table, too. 

\begin{figure}[b]
    \centering
    \begin{subfigure}{0.23\textwidth}
        \includegraphics[width=\textwidth]{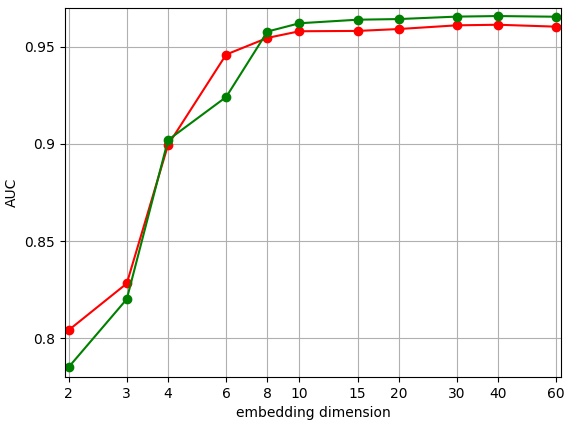}
        \caption{Results for node M4
        \label{fig:sub_a}}
    \end{subfigure}
    \hfill
    \begin{subfigure}{0.23\textwidth}
        \includegraphics[width=\textwidth]{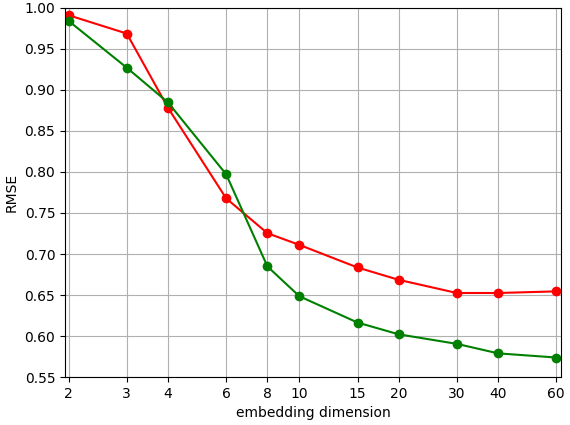}
        \caption{Results for node M6
        \label{fig:sub_b}}
    \end{subfigure}
    \caption{Quality measures as a function of the embedding dimension (logarithmically scaled) when using just the main table (red) and using the combined information of both tables (green).
    \label{fig:results}}
\end{figure}

In order to perform regression or classification tasks for several target columns,
we compute task-independent table entry embeddings
with respect to the EmbDI procedure \cite{embdi2020}, for a variety of embedding dimensions.
First, we consider the main table only. For each training row, we compute a row embedding vector by averaging the
embeddings of all entries in this row, excluding the target columns to be task-independent. This row embedding procedure can be applied not only to training rows but also to test rows, as the entries in the test rows are drawn from the same underlying distribution as the entries in the training rows.
For all target columns, we then perform a regression or classification task, depending on the type of pooling function (numerical as mean, median, norm, ... or categorical). For any test row, we search for 10 nearest neighbors in the row embedding space of the training rows. The prediction is computed as an average of the target values of the selected 10 training rows, weighted by the inverse distance of the test row embedding vector to the 10 nearest training row embedding vectors. 
The quality of the prediction is measured by the root-mean-squared error (RMSE) for numerical targets and by the area under the receiver operating characteristic curve (AUC) for categorical targets. These quality measures for two nodes of the example dataset are shown in red in Fig.~\ref{fig:results} with respect to the embedding dimension. 

Second, we apply the EmbDI embedding procedure to the main and additional table simultaneously. We use all the rows of the additional table for training, while for the main table we reserve 10\% for testing as before. 
Again, regression and classification tasks are conducted for the target columns of the main table. The results for the two selected nodes are highlighted in green in Fig.~\ref{fig:results}. As the merged information requires a higher embedding dimension to be fully represented, the comparison of the two curves in Fig.~\ref{fig:results} is meaningful for sufficiently high embedding dimensions. There, involving the additional table improves the results for targets influenced by latent information from the additional dataset. 

Hence, we showed that our method for synthetic relational dataset generation is able to construct realistic related tables in the sense that the additional table contains information that is not present in the main table, but influences the target columns of the main table. This is an essential, frequently occurring property of real-world relational datasets. We emphasize that further research should include a more comprehensive evaluation with more datasets and further methods for downstream tasks, see Section~\ref{sec:limitations}.

\section{Discussion and Conclusion}
In this work, we presented an approach for generating relational datasets based on SCMs. The corresponding graph controls the causality between features and targets, involving latent causal relationships to model inter-table dependencies. Our approach serves as a scalable methodology to provide huge amount of data with various statistical properties for robust training of a tabular foundation model for relational tabular data.

The main advantage to use SCMs is the ability to model causal relationships. Thus, we are able to control the dependence between certain targets and features. Additionally, by incorporating isolated sub-graphs, we can generate data that is irrelevant to the targets mimicking real-world redundancy. 

The quality of generated data is determined by several parameters. Choosing a large hidden dimension $n$ and projecting the data to a one-dimensional output may significantly increase the difficulty of predicting a target based on the feature nodes. Furthermore, the choice of the activation function of the neural networks influences data complexity.

A key strength of our approach lies in its ability to generate relational tables that capture complex causal relationships, including those mediated by latent variables. This simulation of inter-table dependencies, often lacking in simpler methods, is crucial for developing robust table representation learning models that can effectively handle the complexities of real-world data, commonly encountered in database management systems.

\section{Limitations}\label{sec:limitations}
Our approach successfully generates relational datasets with numerical and categorical features. However, a more detailed experimental analysis with varying parameters, that would go beyond the scope of this short paper, is desirable. Real-world databases often contain multimodal elements like images and text. Extending our framework to incorporate these diverse data types represents an important research direction. Furthermore,
a comprehensive evaluation for three or more relational tables, including cross-connections, needs to be conducted.

\newpage
\bibliography{custom}

\newpage
\appendix

\section{Example of Generated Relational Tables}
\label{sec:appendix}
Based on the graph in Fig.~\ref{fig:combined_graph}, we sample two relational tables. The hidden dimension is set to $n=2$. For the root node M0, the data follows a gamma distribution with shape $\alpha = 2.245$ and scale $\theta=1.780$. The data at root node A0 is drawn from a normal distribution with mean $\mu=-0.029$ and standard deviation $\sigma=0.816$, and for each component of the data vector at root node A2, we choose randomly with $p=0.5$ if the component is drawn from a standard normal or an exponential distribution with scale $\lambda=0.584$. A randomly chosen fraction of 10\% of the data is affected by noise, where the noise standard deviation is set to 0.1. The pre-sampling run described in Subsection~\ref{subsec:data_sampling} is conducted with 1,000 samples. For the categorical nodes, the following numbers of categories are chosen: M0:~6, M4:~2, M7:~6, A0:~3, A1:~4, A2:~2. All these numbers are drawn randomly from a normal distribution with mean $\mu=4$ and standard deviation $\sigma=2$. For the categorical coupling node C, the number of categories is 175, chosen randomly from a normal distribution with mean $\mu=100$ and standard deviation $\sigma=50$, mimicking a foreign key column.

\begin{table}[h]
\centering
\tiny
\begin{tabular}{c c c c c c c c c}
\hline
M0 & M1 & M2 & M3 & M4 & M5 & M6 & M7 & C \\
\hline
4 & -0.831 & 1.281 & 0.669 & 0 & 1.722 & 2.418 & 0 & 53 \\
0 & -0.556 & 1.190 & 0.239 & 0 & 1.630 & 2.204 & 3 & 46 \\
5 & -0.243 & 1.325 & 0.932 & 0 & 1.765 & 2.100 & 0 & 48 \\
2 & -0.627 & 1.276 & 0.927 & 0 & 1.718 & 2.563 & 0 & 46 \\
3 & -0.398 & 1.154 & 0.295 & 0 & 1.591 & 2.020 & 3 & 15 \\
\hline
\end{tabular}
\caption{Main Table for the DAG shown in Fig.~\ref{fig:combined_graph}, the first 5 out of 100,000 rows are displayed.
\label{tab:main_table}}
\end{table}

\begin{table}[h]
\centering
\tiny
\begin{tabular}{c c c c c c}
\hline
A0 & A1 & A2 & A3 & A4 & C \\
\hline
0 & 0 & 1 & 0.499 & 0.355 & 46 \\
0 & 0 & 0 & 1.005 & 2.903 & 76 \\
0 & 3 & 0 & 0.661 & 0.711 & 59 \\
1 & 2 & 0 & 0.567 & 0.577 & 103 \\
2 & 3 & 1 & 0.271 & 0.691 & 146 \\
\hline
\end{tabular}
\caption{Additional Table for the DAG shown in Fig.~\ref{fig:combined_graph}, the first 5 out of 500 rows are displayed.
\label{tab:add_table}}
\end{table}

\end{document}